\documentclass[conference]{IEEEtran}

\usepackage{amsmath,amssymb,amsfonts}
\usepackage{graphicx}
\usepackage{makecell}
\usepackage{textcomp}
\usepackage[svgnames, table]{xcolor}
\usepackage{minted}
\usepackage{multirow}
\usepackage{enumitem}
\usepackage{caption}
\usepackage{subcaption}
\usepackage{hyperref}
\usepackage{setspace}
\usepackage[nocompress]{cite}
\usepackage{balance}

\usepackage[most]{tcolorbox}

\usepackage{tikz}
\usetikzlibrary{positioning,fit,shadows.blur,shapes.misc}

\definecolor{good}{HTML}{2E7D32}   
\definecolor{bad}{HTML}{C62828}    
\definecolor{neutral}{HTML}{1565C0} 

\tikzset{
  panel/.style={rounded corners=4pt, draw=black!60, fill=black!2, inner sep=6pt,
  blur shadow={shadow blur steps=8, shadow blur extra rounding=.4pt}
  },
  title/.style={font=\bfseries, align=center, text=black},
  codebox/.style={draw=black!40, fill=white, rounded corners=3pt, inner sep=5pt, align=left, font=\ttfamily\scriptsize},
}

\newcommand{\goodline}[1]{\textcolor{good}{\checkmark\ #1}}
\newcommand{\badline}[1]{\textcolor{bad}{\(\times\)\ #1}}
\newcommand{\neutline}[1]{\textcolor{neutral}{#1}}
\newcommand{\squishlist}{
  \begin{list}{$\bullet$}
    { \setlength{\itemsep}{0pt}      \setlength{\parsep}{0pt}
      \setlength{\topsep}{0pt}       \setlength{\partopsep}{0pt}
      \setlength{\leftmargin}{1em} \setlength{\labelwidth}{1em}
      \setlength{\labelsep}{0.5em} } }
\newcommand{\squishlisttwo}{
  \begin{list}{$\bullet$}
    { \setlength{\itemsep}{0pt}    \setlength{\parsep}{0pt}
      \setlength{\topsep}{0pt}     \setlength{\partopsep}{0pt}
      \setlength{\leftmargin}{2em} \setlength{\labelwidth}{1.5em}
      \setlength{\labelsep}{0.5em} } }
\newcommand{\squishlistend}{
    \end{list}  }

\begin{document}

\IEEEoverridecommandlockouts
\makeatletter\def\@IEEEpubidpullup{6.5\baselineskip}\makeatother
\IEEEpubid{\parbox{\columnwidth}{
		Network and Distributed System Security (NDSS) Symposium 2027\\
		22 - 26 March 2027 , Seoul, Republic of Korea\\
		ISBN 979-8-9919276-8-0\\  
		https://dx.doi.org/10.14722/ndss.2027.[23$|$24]xxxx\\
		www.ndss-symposium.org
}
\hspace{\columnsep}\makebox[\columnwidth]{}}

%
\title{\Large \bf Beyond Embeddings: Interpretable Feature Extraction for Binary Code Similarity}

\author{\IEEEauthorblockN{Charles Edward Gagnon}
\IEEEauthorblockA{Defence Research and Development Canada\\
McGill University}
\\
\IEEEauthorblockN{Philippe Charland}
\IEEEauthorblockA{Defence Research and Development Canada}
\and
\IEEEauthorblockN{Steven H. H. Ding}
\IEEEauthorblockA{McGill University}
\\
\\
\IEEEauthorblockN{Benjamin C. M. Fung}
\IEEEauthorblockA{McGill University}
}

\maketitle

\begin{abstract}

Binary code similarity detection is a core task in reverse engineering. It
supports malware analysis and vulnerability discovery by identifying
semantically similar code in different contexts. Modern methods have progressed
from manually engineered features to vector representations. Hand-crafted
statistics (e.g., operation ratios) are interpretable, but shallow and fail to
generalize. Embedding-based methods overcome this by learning robust
cross-setting representations, but these representations are opaque vectors that
prevent rapid verification. They also face a scalability--accuracy trade-off, since
high-dimensional nearest-neighbor search requires approximations that reduce
precision. Current approaches thus force a compromise between interpretability,
generalizability, and scalability.
 
We bridge these gaps using a language model-based agent to conduct structured
reasoning analysis of assembly code and generate features such as input/output types, side
effects, notable constants, and algorithmic intent. Unlike hand-crafted
features, they are richer and adaptive. Unlike embeddings, they are
human-readable, maintainable, and directly searchable with inverted or
relational indexes. Without any matching training, our method respectively
achieves 42\% and 62\% for recall@1 in cross-architecture and cross-optimization
tasks, comparable to embedding methods with training (39\% and 34\%). Combined
with embeddings, it significantly outperforms the state-of-the-art,
demonstrating that accuracy, scalability, and interpretability can coexist.

\end{abstract}


%
\IEEEpeerreviewmaketitle

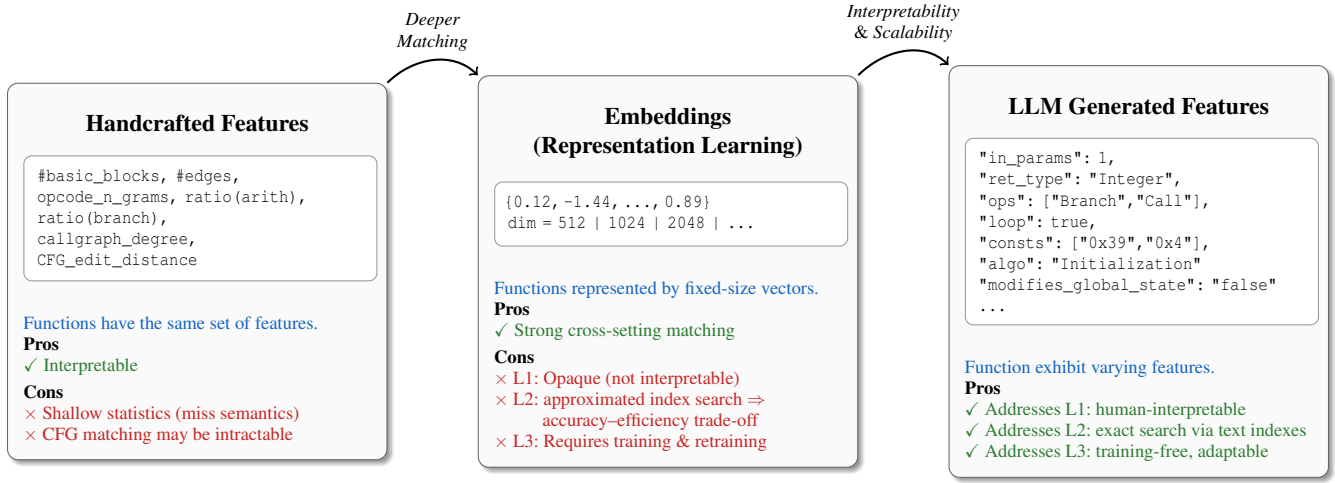
\begin{figure*}[t]
\centering
\begingroup\scriptsize
\begin{tikzpicture}

\node[panel, minimum width=0.22\linewidth] (p1) {
  \begin{minipage}{0.26\linewidth}
  \centering
  {\small \tikz{\node[title]{Handcrafted Features};}}\\[2pt]
  \tikz{\node[codebox, text width=\linewidth-8pt]{
    \#basic\_blocks,\ \#edges,\\
    opcode\_n\_grams,\ ratio(arith), ratio(branch),\\
    callgraph\_degree,\\
    CFG\_edit\_distance
  };}\\[4pt]
  \begin{flushleft}
    \neutline{Functions have the same set of features.} \\
    \textbf{Pros}\\
    \goodline{Interpretable}\\[2pt]
    \textbf{Cons}\\
    \badline{Shallow statistics (miss semantics)}\\
    \badline{CFG matching may be intractable}
  \end{flushleft}
  \end{minipage}
};

\node[panel, right=12mm of p1, minimum width=0.22\linewidth] (p2) {
  \begin{minipage}{0.26\linewidth}
  \centering
  {\small \tikz{\node[title]{Embeddings \\ (Representation Learning)};}}\\[2pt]
  \tikz{\node[codebox, text width=\linewidth-8pt]{
    \{0.12,\,-1.44,\,\dots,\,0.89\}\\
    dim = 512 | 1024 | 2048 | ...
  };}\\[4pt]
  \begin{flushleft}
    \neutline{Functions represented by fixed-size vectors.} \\
    \textbf{Pros}\\
    \goodline{Strong cross-setting matching}\\[2pt]
    \textbf{Cons}\\
    \badline{L1: Opaque (not interpretable)}\\
    \badline{L2: approximated index search $\Rightarrow$ \\ \;\,\,\qquad accuracy--efficiency trade-off}\\
    \badline{L3: Requires training \& retraining}
  \end{flushleft}
  \end{minipage}
};

\node[panel, right=12mm of p2, minimum width=0.22\linewidth] (p3) {
  \begin{minipage}{0.26\linewidth}
  \centering
  {\small \tikz{\node[title]{LLM Generated Features};}}\\[2pt]
  \tikz{\node[codebox, text width=\linewidth-8pt]{
    "in\_params": 1,\newline
    \ \ "ret\_type": "Integer",\newline
    \ \ "ops": ["Branch","Call"],\newline
    \ \ "loop": true,\newline
    \ \ "consts": ["0x39","0x4"],\newline
    \ \ "algo": "Initialization"\newline
    \ \ "modifies\_global\_state": "false"\newline
    \ \ ... 
  };}\\[4pt]
  \begin{flushleft}
    \neutline{Function exhibit varying features.} \\
    \textbf{Pros}\\
    \goodline{Addresses L1: human-interpretable}\\
    \goodline{Addresses L2: exact search via text indexes}\\
    \goodline{Addresses L3: training-free, adaptable}\\
  \end{flushleft}
  \end{minipage}
};

\draw[->, thick]
  (p1.north east) to[out=45, in=135]
  node[above, yshift=1pt, align=center]{\scriptsize\itshape Deeper\\\itshape Matching}
  (p2.north west);
\draw[->, thick]
  (p2.north east) to[out=45, in=135]
  node[above, yshift=1pt, align=center]{\scriptsize\itshape Interpretability\\\& \itshape Scalability}
  (p3.north west);

\end{tikzpicture}
\endgroup
\caption{Evolution of features in BCSD. \textbf{Left:} Handcrafted statistics are interpretable, but shallow and sometimes intractable (CFG). \textbf{Middle:} Deep embeddings improve matching, but suffer from L1--L3 (interpretability, scalability--accuracy, training dependency). \textbf{Right:} Our LLM-generated structured features are interpretable, searchable via inverted/relational indexes, and training-free, addressing L1--L3, while improving practicality.}
\label{fig:motivation-bcsd}
\end{figure*}

\section{Introduction}

Modern software development heavily relies on external libraries.
For security researchers, detecting whether an executable invokes vulnerable library functions is critical for assessing and mitigating exposure~\cite{BCSD, BCSDsurvey}. For reverse engineers, identifying redundant assembly functions accelerates analysis and enables them to focus on the unique logic of a binary. Binary Code Similarity Detection (BCSD) provides a means to address these requirements by identifying whether two compiled code fragments exhibit similar behavior. Its importance is amplified by the growing size, modularity, and production rate of modern software, which make manual inspection infeasible. Beyond vulnerability assessment, BCSD also plays an important role in malware analysis, software supply-chain auditing, and firmware security, where statically linked libraries are common and reused components can be difficult to identify \cite{ERS0,sbom-difficulty}. For instance, when a vulnerability is discovered in a library, BCSD enables efficient identification of affected binaries or firmware, supporting both defensive security and large-scale reverse engineering.

Early approaches to BCSD used human-defined heuristics to extract a “feature vector” from a binary code fragment~\cite{op-seq, BinDiff, clones.net}. These heuristics could be derived statically by examining a function and its control-flow graph (CFG)—for example, by measuring the number of basic blocks, the ratio of arithmetic to control-flow instructions, or the sequence of opcodes—or dynamically by executing the function in an emulator to capture instruction counts, memory access patterns, or system call traces ~\cite{kam1n0}. Such methods were deterministic and had the advantage of producing human-understandable feature vectors, but they suffered key limitations. Purely statistical descriptors were often too simplistic to capture deep semantic similarity~\cite{op-seq}, while more sophisticated CFG-based comparisons sometimes required computationally intractable algorithms such as subgraph isomorphism or graph-edit distance~\cite{BinDiff}, which limited their scalability.

More recently, machine learning (ML)-based approaches have demonstrated superior performance in binary code similarity detection ~\cite{SAFE,PalmTree,OrderMatters,Asm2Vec,CLAP}. Inspired by natural language processing, these methods map each code fragment into a high-dimensional floating-point embedding that captures its structural and semantic patterns. Similarity is then computed with vector distance metrics such as cosine similarity, and large-scale retrieval relies on approximate nearest-neighbor search. These models have shown strong generalization across compilers, optimizations, and architectures.
However, embedding-based approaches suffer from three limitations:

\squishlist
\item 
\textbf{(L1) Lack of interpretability and maintainability.}
The embeddings are opaque vectors with no intuitive meaning. Analysts cannot easily understand why two fragments match, making clone verification difficult and limiting their usefulness in reverse engineering workflows.

\item 
\textbf{(L2) Scalability–accuracy trade-off.}
Embeddings are high-dimensional vectors and searching at scale requires approximate nearest-neighbor indexing. While this improves efficiency, it sacrifices precision, whereas exact search is prohibitively slow—forcing a compromise between scalability and accuracy.

\item 
\textbf{(L3) Dependency on training data.}
These models must be trained on large datasets of code fragments and their effectiveness is constrained by the coverage of the training data (e.g., architectures, compilers, optimization levels). As a result, they often fail to generalize to unseen settings or proprietary instruction sets without costly retraining.
\squishlistend

Our work presents a method to detect code fragment clones across binaries using pre-trained large language models (LLMs), without any training or fine-tuning. Instead of opaque embeddings, our approach generates \textit{structured, human-interpretable feature vectors} (e.g., input/output types, loop structures, high-level behaviors, inferred algorithms), directly addressing \textbf{L1 (lack of interpretability)} by allowing analysts to understand why two fragments match. Because these features are represented as text, they can be indexed and queried using inverted or relational databases, avoiding high-dimensional nearest-neighbor search and thereby overcoming \textbf{L2 (scalability--accuracy trade-off)}. Moreover, by relying on the general reasoning ability of modern LLMs rather than curated training datasets, our method naturally generalizes to \textbf{unseen compilers, optimizations, and architectures}, addressing \textbf{L3 (dependency on training data)}. In this way, our method combines the transparency of early heuristic approaches with the robustness of modern ML, while also surpassing state-of-the-art results (see Figure~\ref{fig:motivation-bcsd}). Our contributions are:  
\squishlist
    \item A training-free BCSD method using LLMs that produces \textit{human-interpretable features} and works across optimizations and architectures.  
    \item A demonstration that performance scales with LLM size, outperforming state-of-the-art models in both accuracy and versatility.  
    \item A hybrid framework that combines our features with embedding-based models, achieving superior results.  
    \item A discussion of efficiency improvements and deployment considerations for large-scale use.  
\squishlistend
The paper is organized as follows: Section~\ref{sec:method} presents our LLM-based feature extraction pipeline; Section~\ref{sec:exp} details the experimental setup and results; Section~\ref{sec:related} reviews related work; and Section~\ref{sec:conclusion} summarizes our contributions and limitations.

\section{From Embedding-Based BCSD to LLM-Based Feature Extraction}
\label{sec:method}

Security researchers and reverse engineers are routinely tasked with the analysis of unknown or proprietary executables.
Reverse engineers try to analyze the binary to understand its underlying algorithms, while security researchers want to assess
the risk associated with potential vulnerabilities found within the executable. This process is usually conducted using
a software reverse engineering tool such as Ghidra ~\cite{ghidra} or IDA Pro ~\cite{ida}. The main functions of these tools are to
disassemble and decompile the provided machine code so that its content can be analyzed by humans. Disassembly is
the process of retrieving the assembly instructions (human-readable machine code) from the binary executable, whereas decompilation
is the process of generating higher-level pseudo-code from the instructions based on common patterns and heuristics.

Binary analysis is a difficult task because once a program is compiled, most of the information contained in its source code
is lost ~\cite{BCSDsurvey}. Variables, data structures, functions, and comments are removed, because the compiler's task is to make
the program as efficient as possible---which often means removing as much as possible. The optimizers within the compiler
only have a single rule: They must not make changes to the observable behavior of the program (often referred
to as the ``as-if rule'' \cite{c++11}). As a result, compilers can remove, reorder, and inline significant parts of the code, making
it difficult to understand the code's behavior. Even worse, adverserial programs such as malware or digital rights management software
make use of obfuscation techniques to resist having their code reverse engineered.

BCSD is the task of determining whether two fragments of binary code perform similar actions.
These fragments are usually first disassembled and are then compared for similarity. In practice,
similarity detection is performed with one known fragment (either because it was analyzed before
or because its source code is known), and one unknown fragment. Known code fragments are typically collected in a
database which is queried against for clone search. If the unknown piece of code is deemed
highly similar to the known one, the analysis task becomes significantly simpler and duplicate work is minimized. For example,
if a vulnerability in a widely used open-source component is found, BCSD can be used to quickly
assess if a binary contains the vulnerable code fragment. It can also be used for plagiarism detection, which
could take the form of license or patent infringement, for software and malware classification ~\cite{op-seq}, or for 
security patch analysis \cite{patch}.

\begin{figure}[t]
\centering
\begin{minipage}[t]{0.5\linewidth}   
  \begin{subfigure}[t]{\linewidth}
    \begin{minted}[fontsize=\scriptsize,baselinestretch=0.9]{nasm}
; x86-64 O0
    push rbp
    mov rbp, rsp
    sub rsp, 8
    mov [rbp+var_8], rdi
    mov rax, [rbp+var_8]
    mov rdi, rax
    call sub_42b49a
    mov rax, [rbp+var_8]
    mov dword ptr [rax+50h], 0
    mov rax, [rbp+var_8]
    mov dword ptr [rax+58h], 0
    mov rax, [rbp+var_8]
    mov edx, [rax+58h]
    mov rax, [rbp+var_8]
    mov [rax+54h], edx
    \end{minted}
  \end{subfigure}
  \vspace{0.5em} 

  \begin{subfigure}[t]{\linewidth}
    \begin{minted}[fontsize=\scriptsize,baselinestretch=0.9]{nasm}
; x86-64 O3
    movdqa xmm0, cs:xmmword_8750
    mov dword ptr [rdi+50h], 0
    mov dword ptr [rdi+58h], 0
    movups xmmword ptr [rdi],xmm0
    mov dword ptr [rdi+54h], 0
    \end{minted}
  \end{subfigure}
\end{minipage}%
\hfill
\begin{minipage}[t]{0.5\linewidth}   
  \begin{subfigure}[t]{\linewidth}
    \begin{minted}[fontsize=\scriptsize,baselinestretch=0.9]{nasm}
; MIPS O0
    addiu $sp, -0x20
    sw $ra, 0x18+var_s4($sp)
    sw $fp, 0x18+var_s0($sp)
    move $fp, $sp
    sw $a0, 0x18+arg_0($fp)
    lw $v0, 0x18+arg_0($fp)
    move $a0, $v0
    jal sub_43D930
    nop 
    lw $v0, 0x18+arg_0($fp)
    sw $zero, 0x50($v0)
    lw $v0, 0x18+arg_0($fp)
    sw $zero, 0x58($v0)
    lw $v0, 0x18+arg_0($fp)
    lw $v1, 0x58($v0)
    lw $v0, 0x18+arg_0($fp)
    sw $v1, 0x54($v0)
    nop 
    move $sp, $fp
    lw $ra, 0x18+var_s4($sp)
    lw $fp, 0x18+var_s0($sp)
    addiu $sp, 0x20
    jr $ra
    \end{minted}
  \end{subfigure}
\end{minipage}

\caption{\texttt{MD5Init} function compiled under different architectures
and optimization levels. BCSD methods aim at identifying them as clones.}
\label{asm-diff}
\end{figure}
\subsection{General BCSD Pipeline}

BCSD is typically organized as a multi-stage pipeline:

\begin{enumerate}
    \item \textbf{Disassembly.} The binary executable is first disassembled into assembly code, using tools such as IDA Pro or Ghidra. In some cases, this step is extended with decompilation to recover higher-level pseudo-code.

    \item \textbf{Preprocessing.} Many approaches transform the raw disassembly into an intermediate representation (IR), CFG, or apply normalization steps (e.g., operand abstraction or instruction reordering). These transformations aim to reduce syntactic variance caused by compilers or obfuscators.

    \item \textbf{Vectorization.} Each assembly function is mapped to a feature vector $\Phi(f)$ that captures its semantics. Different approaches instantiate $\Phi$ in different ways: earlier systems relied on manually engineered features, whereas recent work uses deep neural embeddings.

    \item \textbf{Indexing and Search.} Known functions are stored in a database of feature vectors. Given a query function $f_q$, its vector $\Phi(f_q)$ is compared against the database (e.g., via cosine similarity or nearest-neighbor search) to identify similar code fragments.
\end{enumerate}

This structured pipeline provides a practical way to organize large-scale clone search. However, challenges in binary analysis are amplified in BCSD, since two code fragments that look very different syntactically can still share the same observable behavior.

\subsection{The Vectorization Problem}

The central challenge in BCSD lies in the \emph{vectorization step}. Given a function $f$ represented as a sequence of assembly instructions, the goal is to define a mapping
\[
    \Phi : f \mapsto \mathbf{v} \in \mathbb{R}^d
\]
that produces a vector $\mathbf{v}$ which preserves the semantics of $f$.  
In other words, two functions $f_1$ and $f_2$ that are semantically equivalent (e.g., compiled with different compilers, optimization levels, or instruction sets) should be mapped to vectors that are closer in the embedding space than unrelated functions:
\[
f_1 \equiv f_2 \;\;\Rightarrow\;\;
\begin{array}{l}
\text{sim}(\Phi(f_1), \Phi(f_2)) >
\text{sim}(\Phi(f_1), \Phi(f_3)), \\
\forall f_3 \not\equiv f_1
\end{array}
\]

Here, $\text{sim}(\cdot,\cdot)$ denotes a similarity measure such as cosine similarity.  
In practice, BCSD systems rarely evaluate based on absolute thresholds, but rather by the ranking quality of known matches (e.g., top-$k$ accuracy or mean average precision).
This requirement makes vectorization the key bottleneck in BCSD:
\begin{itemize}
    \item \textbf{Feature design.} Early approaches used manually engineered heuristics (e.g., opcode histograms or CFG metrics), which were interpretable but too simplistic to robustly capture semantics.
    \item \textbf{Learning-based embeddings.} Here, an \emph{embedding} refers to a fixed-length vector representation of a function, typically produced by a neural network, that places semantically similar functions close together in the vector space. Recent work uses neural networks to learn $\Phi$, producing high-dimensional embeddings that yield better accuracy. However, these embeddings are opaque, tightly coupled to the training data, and computationally expensive to use at scale.
\end{itemize}

Thus, the vectorization problem is not only about finding a function-to-vector mapping, but also about balancing interpretability, generalization, and efficiency.

\subsection{Limitations of Embedding Vectorization}

In the introduction, we highlighted three central limitations (L1–L3) of embedding-based approaches. We now elaborate on these challenges, as they frame the design space for our method.

\textbf{L1: Lack of Interpretability.}  
The feature vectors produced by deep learning models are high-dimensional and opaque. When two fragments are judged to be similar, the embedding offers no explanation of what instructions, structures, or semantic properties led to that conclusion. This makes it difficult for analysts to validate results or to trust the system in high-stakes settings such as vulnerability detection. In contrast, traditional heuristic-based BCSD methods generated explicit features such as opcode sequences or control-flow structures, which—while less powerful—at least provided human-understandable reasoning. The black-box nature of embeddings therefore limits their usefulness in workflows where analyst oversight is essential.

\textbf{L2: Scalability of Nearest-Neighbor Search.}  
In production-scale environments, BCSD databases may contain millions of disassembled functions. Each query requires finding the most similar embeddings among these millions of candidates. A naïve search compares the query vector to every database vector, resulting in linear time complexity that becomes prohibitively expensive at scale. To address this, approximate nearest-neighbor (ANN) algorithms are commonly used, which index vectors in specialized data structures (e.g., graphs, trees, or hashing schemes) to accelerate search. However, ANN introduces approximation errors: the true nearest neighbor may be missed, or spurious neighbors may be returned, leading to degraded accuracy. Thus, practitioners face a fundamental trade-off between search efficiency and similarity accuracy.

\textbf{L3: Limited Generalization.}  
Existing embedding-based BCSD models are trained on specific corpora of binaries, often covering only a narrow range of architectures, compilers, and optimization settings. In practice however, reverse engineers must analyze binaries compiled under diverse conditions, often with toolchains and optimization heuristics unseen during training. As a result, these models exhibit poor cross-domain generalization.  
By contrast, human reverse engineers are able to recognize equivalence between functions across compilers or platforms because they rely on general knowledge of instruction semantics and common code generation patterns. Embedding-based methods lack this higher-level reasoning, and thus fail to transfer their knowledge effectively when faced with out-of-distribution inputs.

Taken together, these limitations motivate a different design point: a method that leverages broad reverse engineering knowledge (mitigating L3), produces human-interpretable features (addressing L1), and avoids reliance on approximate vector search (reducing L2). Our approach, based on LLM feature extraction, is explicitly constructed around these principles.

\begin{figure}
\centering

\begin{minipage}[t]{0.55\linewidth}
\vspace{0pt} 
{\linespread{0.9}\selectfont
\begin{minted}[
  fontsize=\scriptsize,
  frame=lines,
  framesep=0.5mm,
  linenos,
  numbersep=0.5pt,
  xleftmargin=0.6em
]{diff}
 { "in_param_cnt": 1,
   "in_param_types": ["Ptr"],
-  "ret_type": "None",
+  "ret_type": "Integer",
   "operation_categories": [
     "ConditionalBranching",
     "SubroutineCall"
   ],
   "loop": false,
   "subcall_targets": 2,
   "indexed_addr": false,
   "int_consts": [
-    "0x39"
+    "0x39","0x4"
   ],
   "float_consts": [],
   "imm_values_cnt": 3,
   "mutates_inputs": false,
   "mutates_globals": false,
   "mem_alloc": false,
   "io_ops": false,
   "block_mem_ops": false,
   "interrupts_syscalls": 0,
-  "algo": "Undetermined",
+  "algo": "Initialization", 
   ...}
\end{minted}
}
\end{minipage}\hfill
\begin{minipage}[t]{0.40\linewidth}
\vspace{0pt} 
\scriptsize
\raggedright
\textbf{Feature explanations}\\[4pt]

\texttt{"in\_param\_types"} — function expects a pointer input parameter.\\[2pt]

\texttt{"loop"} — whether explicit looping constructs are present (here: none).\\[2pt]

\texttt{"subcall\_targets"} — number of distinct callees invoked (here: 2).\\[2pt]

\texttt{"indexed\_addr"} — use of indexed/array-style addressing (false = absent).\\[2pt]

\texttt{"imm\_values\_cnt"} — number of unique immediate constants in code.\\[2pt]

\texttt{"mutates\_globals"} — whether the function writes to global state (false).\\[2pt]

\texttt{"algo"} — high-level role inferred; here = initialization.\\
\end{minipage}

\caption{Comparison of the simplified example \texttt{sha384\_init} assembly function for \texttt{ARM} (red) and \texttt{x86-64} (green); identical values in black. Right-hand panel explains several less obvious features. The specific set of features shown is function-dependent and generated by the language model.}
\label{fig:feature-diff}
\end{figure}

\begin{figure*}
\centerline{\includegraphics[width=\linewidth]{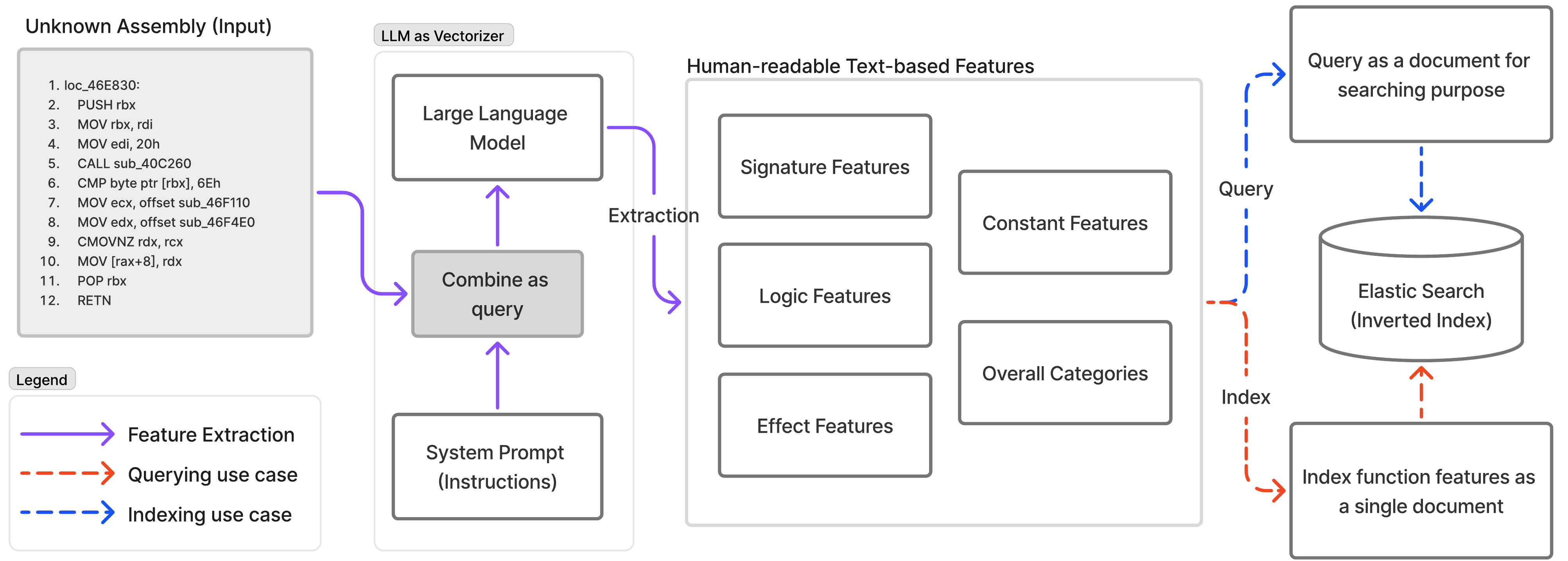}}
\caption{Workflow of our method in both the indexing and searching use cases.}
\label{fig:BCSD-workflow}
\end{figure*}

\subsection{LLM-Based Feature Extraction}

Our design is guided directly by the limitations of embedding-based BCSD identified in the previous section.  
We aim to replace opaque, training-dependent vector embeddings with structured, human-interpretable features 
that can be generated from arbitrary assembly code. This choice is motivated by three design principles:

\squishlist

    \item \textbf{Interpretability (addresses L1).}  
    Our method extracts explicit features—such as input/output types, control structures, side effects, 
    and notable constants—that analysts can directly inspect. When a similarity match is reported, 
    the underlying reasoning can be understood and verified by a human, something impossible with 
    black-box embeddings. See the simplified example in Figure~\ref{fig:feature-diff},
    which illustrates how interpretable features can be directly compared.
    
    \item \textbf{Scalability (addresses L2).}  
    Because features are textual and structured, they can be stored and queried using inverted indexes 
    or relational databases. This eliminates the need for approximate nearest-neighbor search in high-dimensional spaces, removing the trade-off between efficiency and accuracy. 
    
    \item \textbf{Generalization (addresses L3).}  
    Instead of training models on narrow corpora of binaries, we leverage large language models that are 
    pretrained on broad, diverse data. These models encode general knowledge of program structure and 
    algorithmic behavior, which transfers across architectures, compilers, and optimization settings.  
\squishlistend

These designs define a different point in the BCSD design space: instead of embedding functions 
into opaque vectors, we treat the LLM as a semantic feature extractor, producing generalizable, interpretable, and scalable representation.

Figure~\ref{fig:BCSD-workflow} illustrates our workflow.  
Given the raw assembly code of a function, the LLM produces a JSON object containing human-readable attributes such as type signatures, control-flow indicators, 
constants, side effects, and overall functional categories. Each function is thus represented as a lightweight document 
rather than a dense vector. These documents are stored in a database that can be indexed using inverted indexes or 
relational queries, enabling efficient and exact similarity search at scale.  

At query time, the process is identical: the unknown function is analyzed by the LLM, its features are generated in the same format, 
and similarity is computed over the textual database. This workflow eliminates approximate nearest-neighbor search and allows 
analysts to directly inspect the reasons behind a reported match using simple diffing on features. 

\subsection{Feature Extraction via Prompting}

The core of our approach is a structured prompt that instructs the LLM to act as a reverse engineering assistant and to output its analysis as a JSON object. Unlike prior methods, our pipeline does not require preprocessing: raw assembly code is fed directly 
into the model, which produces a structured set of semantic features. The full prompt is available as part of the published 
artifacts ~\cite{artifact}.  

\paragraph{Framing and Conditioning} We use a system prompt that defines the task, the expected format, and the role of the LLM. This preamble ensures consistency 
and constrains the model output. An excerpt is shown below:

\begin{tcolorbox}[colback=gray!10,colframe=gray!20,
 coltitle=black,]
\small
You are an expert assembly code analyst, specializing in high-level semantic description and feature extraction for comparative
analysis. Your goal is to analyze an assembly function from an unspecified architecture and compiler and provide its extracted
high-level features, formatted as a \texttt{JSON} object. For the provided assembly routine, extract the following features and infer the
algorithm. Your output \textbf{MUST} be a \texttt{JSON} object conforming to the structure defined by these features.
\end{tcolorbox}

\paragraph{Type Signature}
The type signature is the foundation of understanding a function’s interface: how many conceptual inputs it takes, whether they are 
raw integers or pointers to memory regions, and what kind of return value is produced.  
This mirrors what reverse engineers attempt manually when labeling functions in tools such as IDA or Ghidra.  

Because compiler transformations and source-level variants may alter the calling convention (e.g., passing values through registers, 
folding constants, or introducing hidden parameters), we define these as \emph{effective parameters}: the distinct conceptual 
inputs that the function’s logic depends on, rather than the literal register or stack arguments.  
This abstraction helps align variants of the same routine that may differ syntactically but share the same semantics.  The LLM is instructed to output the following fields:

\squishlist
    \item \emph{Input Parameter Count (Integer):}  
    The number of effective conceptual inputs used by the function. This can be inferred by observing how values are consumed 
    (e.g., registers initialized at entry, stack offsets consistently read from, or constants passed into subroutines).  
    The goal is not ABI-level accuracy but a stable abstraction across compiler and optimization variants.  

    \item \emph{Input Parameter Types (Array of Strings):}  
    Each inferred input is labeled as either \emph{Integer} (used in arithmetic, comparisons, or scalar logic) 
    or \emph{Pointer} (dereferenced or used as an address). Ambiguous cases (e.g., integers occasionally cast to addresses) 
    are resolved by majority usage within the function body. We observed that inferring more information than these two primitive types has shown to be too complicated for current LLMs. 

    \item \emph{Return Value Type (String):}  
    The abstract type of the returned value, if any. This is usually inferred from the value placed in the conventional return register 
    (e.g., \emph{EAX}/\emph{RAX}) or stack slot before a \emph{RET}. If no consistent return is found, the type is set to \emph{None}.  
\squishlistend

\paragraph{Logic and Operations}

While the type signature describes the function’s interface, its internal logic reveals how it behaves algorithmically.  
For reverse engineers, identifying loops, dispatch tables, or the prevalence of arithmetic versus memory operations 
is often the fastest way to infer purpose. Because compiler optimizations alter instruction order, 
we capture these features at the level of semantic patterns rather than raw mnemonics.  
The LLM is instructed to output the following fields:
\squishlist
    \item \emph{Loop Indicators (Boolean):}  
    True if the function contains backward branches (conditional jumps targeting earlier instructions) 
    or explicit loop constructs, which imply iterative structure. False otherwise.  

    \item \emph{Jump Table Indicators (Boolean):}  
    True if the function implements control transfer via jump tables, inferred from indirect jumps based on 
    computed indices or long sequences of compare-and-jump instructions.  

    \item \emph{Use of Indexed Addressing Modes (Boolean):}  
    True if the function accesses memory via complex addressing forms such as [base + index * scale + offset].  
    This usually indicates array traversals, buffer processing, or structured data access.  

    \item \emph{Presence of SIMD Instructions (Boolean):}  
    True if the function employs Single Instruction, Multiple Data (SIMD) instructions or wide registers.  
    This often signals numerical routines, multimedia processing, or cryptographic kernels.  

    \item \emph{Number of Distinct Subroutine Call Targets (Integer):}  
    The number of unique functions invoked. A high value suggests orchestration routines, 
    whereas a value of zero marks leaf utility functions. Inline calls eliminated by the compiler are not counted.  

    \item \emph{Dominant Operation Categories (Array of Strings):}  
    A coarse-grained classification of the function’s logic, chosen from: arithmetic, bitwise, data movement, 
    conditional branching, subroutine call, or memory access.  
    This mirrors how reverse engineers mentally classify functions — for example, a hashing routine may combine 
    bitwise and arithmetic, while a dispatcher may emphasize control flow.  
\squishlistend

\paragraph{Notable Constants}
Constants embedded in a function often serve as signatures of higher-level algorithms.  
Cryptographic routines can be identified by well-known primes, irreducible polynomials, or fixed S-box values;  
file format parsers by magic numbers at fixed offsets; and network code by protocol constants or port numbers.  
For reverse engineers, spotting such values is often the quickest way to narrow down a function’s purpose.  

Earlier approaches typically relied on regular expressions or literal extraction to gather constants from assembly.  
This produced large sets dominated by trivial values (loop counters, stack offsets, small increments), requiring 
manual filtering to isolate the meaningful ones. By contrast, a language model can act more like an experienced 
analyst: it not only identifies constants but also distinguishes trivial from informative values, and normalizes them 
into a consistent representation. This enables database-level comparison without ad hoc filtering.  
The language model is instructed to extract:
\squishlist
    \item \emph{Presence of Notable Integer Constants (Array of Hexadecimal Strings):}  
    A list of up to 15 unique integer literals represented in hexadecimal form. Trivial values such as 0, 1, -1, 
    or common stack adjustments excluded, while unusual or patterned constants (e.g., 0x04C11DB7, 0xDEADBEEF) 
    are preserved.  

    \item \emph{Count of Distinct Immediate Values (Integer):}  
    The total number of unique literal values used in the function. This provides a quantitative measure 
    of how “constant-heavy” the routine is, helping to separate lightweight utilities from structured 
    parsers or cryptographic kernels.  

    \item \emph{String Literal Presence (Boolean):}  
    True if identifiable string literals are referenced or manipulated within the function, such as error messages, 
    protocol headers, or format strings. False otherwise. This feature often highlights I/O or interface code.  
\squishlistend

\paragraph{Side Effects}
Beyond inputs and outputs, many functions affect the surrounding program state.  
These effects are especially important for reverse engineers, since they help distinguish 
pure computational routines from those that interact with memory, I/O, or the operating system.  
Traditional static analysis often struggles here: heuristics based on raw instruction matching 
(e.g., scanning for \texttt{INT} or \texttt{SYSCALL}) can detect obvious system calls, 
but fail to capture subtler patterns such as input-pointer modification or implicit error handling.  
An LLM, by contrast, can integrate contextual cues to decide whether a write operation 
is best interpreted as modifying a parameter, updating global state, or performing memory allocation. The language model is instructed to extract:
\squishlist
    \item \emph{Modifies Input Parameters (Boolean):}  
    True if writes target memory derived from input arguments, such as dereferenced pointers passed into the function.  
    False if inputs are only read or used in calculations.  

    \item \emph{Modifies Global State (Boolean):}  
    True if writes occur to fixed or absolute memory addresses, or via pointers likely originating from global data segments.  
    False otherwise. This captures stateful routines that affect program-wide variables.  

    \item \emph{Memory Allocation or Deallocation (Boolean):}  
    True if the function follows patterns of dynamic memory management, such as calls returning usable pointers or frees.  
    Heuristics include detecting arguments consistent with sizes, followed by pointer-based writes.  

    \item \emph{I/O Operations (Boolean):}  
    True if the function appears to perform input/output, such as passing string literals or buffers to read/write calls.  
    False otherwise. This feature often distinguishes utility code from user-facing or system-interfacing routines.  

    \item \emph{Block Memory Operations (Boolean):}  
    True if patterns of copying or setting large memory blocks are observed, either through loops with indexed accesses 
    or calls to known block operations such as \texttt{memcpy}. False otherwise.  

    \item \emph{Error Handling (Boolean):}  
    True if extensive conditional checks occur after subroutine calls, especially comparisons against error codes, 
    followed by logging or cleanup routines. False otherwise.  

    \item \emph{Number of System Calls or Software Interrupts (Integer):}  
    The total number of instructions that directly invoke kernel-mode transitions (e.g., INT, SYSCALL, SVC).  
    This provides a direct measure of OS interaction.  
\squishlistend

\paragraph{Categorization}
While individual features describe specific aspects of a function, reverse engineers also 
seek a concise, high-level label that captures the routine’s overall purpose.  
Such categorization helps analysts prioritize what to study, and it provides an immediate 
explanation for why two functions might be considered similar.  
Earlier BCSD approaches rarely attempted categorization, since heuristics were too brittle 
and embeddings too opaque. With an LLM, however, it is feasible to synthesize a judgment 
based on the combination of signature, operations, constants, and side effects, much like 
a human analyst would.  
The language model is instructed to extract:
\squishlist
    \item \emph{Inferred Category (String):}  
    A high-level label summarizing the function’s purpose, chosen from categories such as:  
    system/OS interaction, memory management, data processing or transformation, control-flow or dispatch,  
    initialization, error handling, utility/helper, cryptographic or hashing, interfacing/wrapper, or undetermined.  
    The LLM bases this decision on the complete set of extracted features, prioritizing consistency and interpretability.  
\squishlistend

\subsection{Generation Robustness}
We adopt several techniques to ensure that the output generated by the LLM is valid JSON and maximally relevant.  
Because LLMs are stochastic generators rather than deterministic parsers, robustness is critical for reliable use at scale.  

\emph{Model performance and scaling.}  
It is well known that model size is a dominant factor in language model quality~\cite{scaling-laws}.  
In our early experiments, smaller local models (e.g., 0.5B parameters) sometimes produced nonsensical output, 
such as repeating the same line until the context was exhausted, or emitting malformed JSON.  
Larger models rarely exhibited such issues, but we designed our pipeline to be robust even under weaker conditions.  

\emph{Retry on failure.}  
To handle invalid JSON, we implement a retry loop: if parsing fails, the same query is repeated with a slightly higher 
sampling temperature. This discourages deterministic repetition of invalid output. In practice, a maximum of three retries 
was sufficient, and over 95\% of responses from the smallest evaluated model were valid on the first attempt.  

\emph{Few-shot prompting.}  
We stabilize the model’s behavior using few-shot prompting~\cite{few-shot}, where several assembly functions with handcrafted 
feature analyses are included in the prompt. In our evaluations, three examples shown to be sufficient: adding more yielded no 
substantial improvement, while using only one still produced reasonable results.  
The examples are deliberately diverse in size, architecture, and logic to reflect the range of our test data. They can be found in our code repository ~\cite{artifact}.

\emph{Schema independence.}  
Commercial LLMs increasingly support schema-enforced decoding (e.g., JSON mode).  
For fairness, we did not use these vendor-specific features, ensuring that results are directly comparable 
to open-source local models that lack such capabilities. Our robustness instead derives purely from prompt design and 
the retry mechanism.  

\begin{table*}
\centering
\resizebox{\linewidth}{!}{
\begin{tabular}{l|r|c|l}
\hline
\hline
Library       & Function count & Median function length & Description                                                          \\ \hline
BusyBox       & 66,706          & 32                     & A fairly complete Unix utility set for any small or embedded system. \\
GNU coreutils & 42,545          & 31                     & The basic utilities of the GNU operating system.                     \\
curl          & 27,198          & 14                     & Command line tool to transfer data with URLs.                        \\
ImageMagick   & 69,873          & 55                     & Software suite used for editing and manipulating digital images.     \\
OpenSSL       & 134,359         & 29                     & Toolkit for general-purpose cryptography and secure communication.   \\
PuTTY         & 12,655          & 26                     & An implementation of SSH and Telnet for Windows and Unix platforms.  \\
SQLite        & 30,322          & 40                     & A small SQL database engine.                                         \\ \hline \hline
\end{tabular}
}
\caption{Detailed statistics of the dataset: function counts, median function lengths, 
and descriptions of the selected software libraries.}
\end{table*}

\subsection{Similarity Computation and Extensibility}
Unlike embeddings, which require vector distances and approximate nearest-neighbor search, 
our structured representation enables direct set-based comparison.  
Each JSON output is flattened into field–value pairs, and similarity is computed using Jaccard overlap.  
This can be implemented efficiently with inverted-index search engines such as Elasticsearch or Lucene, 
allowing clone retrieval to scale without custom ANN infrastructure.  
The schema-based design also provides extensibility: new features can be added incrementally 
(with defaults or derived values) without retraining models or regenerating the database.  
This makes the system both scalable and maintainable over time.

\section{Experiments}
\label{sec:exp}

The evaluation is structured to progressively validate different aspects of our approach.  
We begin by benchmarking a range of candidate LLMs to identify which are most suitable for feature extraction in BCSD.  
Next, we compare our method directly against state-of-the-art machine learning approaches,  
measuring cross-optimization and cross-architecture performance on our dataset.  
To better understand where our method’s strengths come from, we then conduct ablation studies,  
examining the role of model size, the number of few-shot examples, and the contribution of different prompt sections.  
Finally, we test composability: showing that our interpretable features can be combined with generic embedding models,  
and that this hybrid design yields results that significantly surpass existing methods.  
All experiments are run on a server equipped with 8 Intel Xeon Gold 5218 CPU cores, 100 GB of RAM,  
and four NVIDIA Quadro RTX 6000 GPUs with 24 GB of RAM each.

\textbf{Dataset:}
The dataset is made of a varied set of executables, so as to be representative of the diversity found in real world software. It is composed of seven open source binaries:
BusyBox ~\cite{busybox}, GNU coreutils ~\cite{coreutils}, curl ~\cite{curl}, ImageMagick~\cite{image-magick}, OpenSSL ~\cite{openssl},
PuTTY ~\cite{putty}, and SQLite ~\cite{sqlite}. All have permissive licenses that allow their use in our evaluations. 
All executables were compiled with \texttt{gcc} for the following architectures: \texttt{x86-64} (also known as \texttt{AMD64}), \texttt{ARM}, \texttt{MIPS}, and \texttt{PowerPC}.
For each architecture, binaries were generated for all optimization levels (\texttt{O0} to \texttt{O3}) and stripped of debug symbols.
The compiled binaries were dissassembled using IDA Pro ~\cite{ida} and separated into individual functions, yielding a total of \(383,658\) assembly functions.
Functions consisting of less than three instructions were not included as part of the dataset, because of their insignificance.

Pairs of equivalent functions compiled for the same platform but with different optimization levels were made for cross optimization
evaluation. Pairs compiled with the same optimization level but for different platforms were also generated for cross
platform evaluation. For example, in  \autoref{asm-diff}, the variants of \texttt{MD5Init} compiled with \texttt{-O0} (top-left) and
\texttt{-O3} (bottom-left) for \texttt{x86-64} form a pair for cross optimization retrieval, whereas the variants compiled with \texttt{-O0}
for \texttt{x86-64} (top-left) and \texttt{MIPS} (right) form a pair for cross platform retrieval.

\textbf{Evaluation method:}
The mean reciprocal rank (MRR) and first position recall (Recall@1) metrics are used for evaluation and for comparison to other methods
~\cite{deprio,code-not-lang,Asm2Vec,CLAP,SAFE}, following the setup in prior work. 
Specifically, a pool of assembly function pairs is used, where both assembly fragments in a pair come from the same source function. 
For each pair, we compare the generated features for the first element against all second elements of the pairs contained in the pool.

For example, consider a pool of ten pairs \((a_i, b_i)\) for \(i \in [1, 10]\), where \(a_i\) is compiled for the \texttt{ARM}
architecture with optimization level \(3\), and \(b_i\) is compiled for the \texttt{MIPS} architecture with the same optimization
level. The feature set generated for function \(a_1\) is compared for similarity with the features of each \(b_i\) for \(i \in [1, 10]\).
A ranking is generated by ordering these comparisons from most to least similar. For a given pair \((a_1, b_1)\), Recall@1 is 
successful if \(b_1\) is the most similar function to \(a_1\). The reciprocal rank is defined as \(1 / \text{rank}(b_1)\), and the 
MRR metric is obtained by averaging this value across all pairs. Recall@1 is computed as the fraction of successful recalls over the total number of pairs.

\subsection{Pilot Testing for LLM Selection}

Before evaluating our method against the baselines with full dataset, we compare different LLMs available in small scale to select our backbones for the remaining experiments, given the potentially high computational cost.
For the local model, three options are considered: Qwen2.5 Coder ~\cite{qwen2} with sizes \(0.5\)B, \(1.5\)B, \(3\)B, and \(7\)B;
Gemma 3 ~\cite{gemma3} with sizes \(1\)B and \(4\)B; and  Qwen3 \(4\)B ~\cite{qwen3}. We preselected these models because they are open source,
small enough to fit in most modern GPUs when quantized, and small enough to fit within our GPU with \(24\) GB of VRAM unquantized. We did not
consider mixture-of-experts models, because they are much larger and the specificity of our
workload is likely to result in only a subset of experts being heavily used. We selected a small set of tasks that are representative
of the extensive experiments conducted against the baselines. There are two cross optimization evaluations and one cross architecture 
evaluation.

In all experiments on local models, the input context size is limited to \(4,096\) tokens and the maximum output tokens to generate is set to \(512\).
Very large assembly functions that do not fit within the input tokens are truncated. The more recent Qwen3 and Gemma 3
models perform better for their size than Qwen2.5 Coder. They match or surpass it in most metrics while being around \(40\%\)
smaller. The three models perform inference in a similar time frame. On one of our GPUs, Qwen2.5 7B takes 5 seconds on average to
generate an analysis for one assembly function. For smaller models such as Qwen2.5 0.5B and 1.5B, inference is faster since multiple queries
can be batched on the same GPU (because of lower memory consumption). 
We selected Qwen2.5 Coder for the remaining experiments because it provides many smaller size configurations for our ablation study, and is more stable than the newer Gemma 3 and Qwen3 models.

For the commercial model, we preselected GPT 4.1 Mini \cite{gpt4} and Gemini 2.5 Flash \cite{gemini2.5}.
These were chosen mainly because of their low cost and availability. The same subset of evaluations as for local models is performed to determine
the model to use for the remaining evaluations.

To provide a comparable environment to local models, the context window is also truncated to 4096 tokens, even if commercial models
support much larger context sizes.
We selected Gemini 2.5 Flash because it performs best and offers a better infrastructure. Compared to GPT 4.1 Mini's \(10\) seconds
latency per request, Gemini is almost able to handle a request every second, making it easier to iterate on our evaluations.

\subsection{Cost Analysis}

The small context size and low output token usage of this technique makes it economically viable. We consider the example of generating analyses for \(100,000\) feature vectors. In the worst case, we estimate the processing involved to consume \(400\) million input tokens, and generate \(50\) million output tokens. On a single local model instance, our setup is able to generate more than \(16,000\) analyses per day. As such, the full database would be generated within a week. With a commercial model such as Gemini 2.5 Flash, the compute cost would amount to \( 135\$ \). With APIs allowing for batch queries of up to 500 million tokens, a single query is necessary to generate analyses for all \(100,000\) functions. Unlike embeddings, these feature vectors are model agnostic, making their generation a one-time cost.

\subsection{Cross-Optimization Evaluation}

{
    \renewcommand{\arraystretch}{1.1}
    \begin{table}
    \centering
    \begin{tabular}{l|ccc} \hline \hline
    Model            & \tt O0--O3 & \tt O2--O3 & \tt ARM--x86-64 \\ \hline
    Qwen2.5 Coder 7B & 0.558      & 0.725      & 0.498            \\
    Qwen3 4B         & 0.550      & \bf 0.850  & 0.564            \\
    Gemma 3 4B       & \bf 0.571  & 0.839      & \bf 0.594        \\ \hline \hline
    \end{tabular}
    \caption{MRR results of the selected evaluations for local model selection, using a pool size of \(100\).}
    \end{table}
}

This experiment benchmarks the capability of the baselines and our method for the detection of similar code fragments across
different optimization levels. There are five baselines, presented in the same order as in ~\autoref{x-arch}.

\squishlist
\item
\emph{Order Matters}~\cite{OrderMatters} combines BERT~\cite{BERT} with control flow graph embeddings, producing function-level representations via a multi-layer perceptron. It supports cross-optimization and cross-architecture retrieval, though training was limited to \texttt{x86-64} and \texttt{ARM}.

\item
\emph{SAFE}~\cite{SAFE} encodes instructions with word2vec~\cite{word2vec} and aggregates them using a self-attentive network~\cite{SANN}. Like other pre-trained models, it supports cross-architecture similarity but was only trained on \texttt{x86-64} and \texttt{ARM}.

\item
\emph{Asm2Vec}~\cite{Asm2Vec}, an early NLP-based approach, samples instruction sequences from CFG traversals and applies PV-DM~\cite{PV-DM} to embed functions. It inspired later models such as SAFE.

\item
\emph{PalmTree}~\cite{PalmTree}, a BERT-based model, tokenizes instructions and is trained on masked language modeling, context prediction, and def-use prediction. It supports cross-optimization and cross-architecture tasks, though the reference implementation targets compiler-level similarity.

\item
\emph{CLAP}~\cite{CLAP} builds on RoBERTa~\cite{RoBERTa}, directly embedding functions and supporting text--assembly alignment for classification. It was trained on \texttt{x86-64}/\texttt{gcc}, but generalizes to other settings.
\squishlistend

{
    \renewcommand{\arraystretch}{1.1}
    \begin{table}
    \centering
    \begin{tabular}{l|ccc} \hline \hline
    Model            & \tt O0--O3 & \tt O2--O3 & \tt ARM--x86-64 \\ \hline
    Gemini 2.5 Flash & \bf 0.674  & \bf 0.865  & \bf 0.766        \\
    GPT 4.1 Mini     & 0.662      & 0.811      & 0.755            \\ \hline \hline
    \end{tabular}
    \caption{MRR results of the selected evaluations for commercial model selection, using a pool size of \(100\).}
    \end{table}
}

{
    \renewcommand{\arraystretch}{1.3}
    \begin{table*}
    \centering
    \resizebox{\linewidth}{!}{
    \begin{tabular}{l|cccc|cccc}
    \hline \hline
    \multirow{2}{*}{Model} & \multicolumn{4}{c|}{MRR}                                 & \multicolumn{4}{c}{Recall @ 1}                           \\ \cline{2-9}
                           & \tt ARM--x86-64 & \tt PowerPC--x86-64 & \tt MIPS--x86-64 & average   & \tt ARM--x86-64 & \tt PowerPC--x86-64 & \tt MIPS--x86-64 & average   \\ \hline
    Order Matters          & 0.007           & 0.007               & 0.007            & 0.007     & 0.002           & 0.000               & 0.001            & 0.001     \\
    SAFE                   & 0.239           & 0.187               & 0.196            & 0.207     & 0.081           & 0.059               & 0.064            & 0.068     \\
    PalmTree               & 0.037           & 0.036               & 0.018            & 0.030     & 0.031           & 0.013               & 0.007            & 0.017     \\
    Asm2Vec                & 0.242           & 0.293               & 0.417            & 0.317     & 0.085           & 0.113               & 0.231            & 0.143     \\
    CLAP                   & 0.416           & \bf 0.523           & 0.494            & 0.478     & 0.334           & \bf 0.443           & 0.415            & 0.397     \\ \hline
    Qwen 2.5 7B            & 0.263           & 0.201               & 0.202            & 0.222     & 0.165           & 0.108               & 0.110            & 0.128     \\
    Gemini 2.5 Flash       & \bf 0.548       & 0.520               & \bf 0.525        & \bf 0.531 & \bf 0.436       & 0.414               & \bf 0.417        & \bf 0.422 \\ \hline \hline
    \end{tabular}
    }
    \caption{Evaluation of the baselines and our method on cross architecture retrieval with a pool size of \(1,000\).
    All functions are compiled with optimization level \(2\) using \texttt{gcc} for the architectures specified in each column.}
    \label{x-arch}
    \end{table*}
}

\noindent \textbf{Results.} We present the results of the baselines and our method evaluated on both the
largest local model, Qwen2.5-Coder 7B, and the commercially deployed model, Gemini 2.5 Flash.
As evident in \autoref{x-opt}, the hardest retrieval task is between optimization levels \(0\) and \(3\),
highlighting the substantial difference between unoptimized and maximally optimized code (see also \autoref{asm-diff}).
At optimization level \(0\), functions perform a lot of unnecessary actions, such as extensively moving
data between registers and performing conditional evaluation of expressions that return a constant value. The generated code
is mostly left untouched by the optimizer. At optimization level \(3\), the compiler will inline simple functions into the
body of the caller, meaning that jumps and calls to other places in the binary are replaced by the destination's instructions.
Some loops are unrolled, so that each iteration of the loop is laid out sequentially instead of performing a conditional check
and a jump to the loop's initial instruction. Also, instructions can be heavily reordered to achieve best performance for the targeted
hardware, while keeping the observable behavior of the program untouched.

For the most part, the baselines perform worse than expected on the evaluations. As our own dataset is used
rather than the one each baseline is pre-trained on, we believe overfitting to be the cause of this poor performance,
as also observed by Marcelli et al. \cite{cisco}. Compared to the baselines, one of our method's advantage is that it requires no
specific training other than the general pre-training process performed by the model developers. As such, our method is less likely to overfit the dataset. This is seen through our its stability in \autoref{x-opt}: our method consistently performs well, compared to some of the baselines that perform excellently on some tasks, but poorly on others.

\subsection{Cross-Architecture Evaluation}

Different CPU architectures have varying assembly code languages. It is hard for BCSD methods that analyze assembly code to support
multiple architectures. These methods need to accurately represent two functions with completely different syntaxes but with identical
semantics as being very similar in terms of their feature set or embedding. Hence, methods that use CFG analysis have a better chance
at supporting many architectures, since the structure of the CFG itself is architecture agnostic. However, the basic blocks that constitute
this graph are still in assembly code, which does not fully resolve the issue. Furthermore, there exists many different variants of each
instruction set, because each new version of an architecture brings new instructions to understand and support. With deep learning methods,
this requires training or fine-tuning the model to understand a new language variant every time. Afterwards, all embeddings in a BCSD database need to
be regenerated. Our method does not directly address this issue, but brings a significant improvement. It indirectly benefits from the vast
amount of data used to train foundational LLMs. Since a LLM has extensively seen all of the mainstream CPU architectures and their dialects
in its training data, it is able to grasp their meaning and extract features from them. Also, if the model in use seems to poorly comprehend a
specific architecture, it can be replaced with one that better performs the specific platform without invalidating the BCSD database. As seen in ~\autoref{x-arch}, our method slightly surpasses the baselines, but there is still room for improvement.

\subsection{Cross-Obfuscation Evaluation}

Obfuscation techniques drastically alter the resulting assembly code. 
Cross-obfuscation evaluation is thus a very good metric to assess whether or
not the BCSD method is able to truly reason about the semantics of the code, rather
than the syntax. The obfuscation methods used for this evaluation are:
\begin{itemize}
    \item Bogus control flow: Injects unused execution paths containing duplicated
    or junk code.
    \item Control flow flattening: Transforms control flow logic into a single large ``switch'' statement.
    \item Instruction substitution: Replaces standard machine instructions with more complex alternatives. For example,
    modifying the \texttt{NOT} instruction by a series of \texttt{XOR} and \texttt{AND} instructions.
\end{itemize}
As seen in ~\autoref{x-obf}, our method once again performs very well.

{
    \renewcommand{\arraystretch}{1.3}
    \begin{table*}
    \centering
\resizebox{\linewidth}{!}{
\begin{tabular}{l|cccc|cccc}
\hline \hline
\multicolumn{1}{l|}{
\multirow{2}{*}{Model}} & \multicolumn{4}{c|}{MRR}                                                                                       & \multicolumn{4}{c}{Recall @ 1}                                                                                \\ \cline{2-9}
\multicolumn{1}{c|}{}                       & \texttt{None--All}                 & \texttt{None--BCF}                 & \texttt{SUB--FLA}                   & average                    & \texttt{None--All}                 & \texttt{None--BCF}                 & \texttt{SUB--FLA}                   & average                   \\ \cline{1-9}
OrderMatters                                & 0.008          & 0.006          & 0.007          & 0.007          & 0.001          & 0.001          & 0.001          & 0.001          \\
SAFE                                        & 0.256          & 0.181          & 0.264          & 0.234          & 0.0625         & 0.0625         & 0.125          & 0.083          \\
PalmTree                                    & 0.122          & 0.289          & 0.215          & 0.209          & 0.060          & 0.200          & 0.083          & 0.114          \\
Asm2Vec                                     & 0.200          & 0.181          & 0.264          & 0.215          & 0.069          & 0.063          & 0.125          & 0.086          \\
CLAP                                        & 0.583          & \textbf{0.766} & 0.576          & \textbf{0.642} & 0.478          & \textbf{0.652} & 0.465          & 0.532          \\ \cline{1-9}
Qwen 2.5 7B                                 & 0.252          & 0.313          & 0.281          & 0.282          & 0.147          & 0.227          & 0.191          & 0.188          \\
Gemini 2.5 Flash                            & \textbf{0.621} & 0.671          & \textbf{0.630} & 0.640          & \textbf{0.502} & 0.609          & \textbf{0.524} & \textbf{0.545} \\ \hline \hline
\end{tabular}
}
    \caption{Evaluation of the baselines and our method on cross obfuscation retrieval with a pool size of \(1,000\). All functions are compiled for \texttt{x86\_64} using \texttt{clang} with optimization level 2. The obfuscation methods used are bogus control flow (\texttt{BCF}), instruction substitution (\texttt{SUB}), and control flow flattening (\texttt{FLA}).}
    \label{x-obf}
    \end{table*}
}

\subsection{Ablation Testing on LLM Size}

{
    \renewcommand{\arraystretch}{1.3}
    \begin{table*}
    \centering
    \resizebox{\linewidth}{!}{
    \begin{tabular}{l|cccccc|cccccc}
    \hline
    \hline
    \multirow{2}{*}{Model} & \multicolumn{6}{c|}{MRR}                                  & \multicolumn{6}{c}{Recall @ 1}                                                        \\ \cline{2-13}
                           & \tt O0--O1 & \tt O0--O2 & \tt O0--O3 & \tt O1--O3 & \tt O2--O3 & average     & \tt O0--O1 & \tt O0--O2 & \tt O0--O3 & \tt O1--O3 & \tt O2--O3 & average   \\ \hline
    Order Matters          & 0.006      & 0.008      & 0.006      & 0.006      & 0.006      & 0.006       & 0.001      & 0.002      & 0.001      & 0.000      & 0.001      & 0.001     \\
    SAFE                   & 0.189      & 0.200      & 0.189      & 0.218      & 0.171      & 0.193       & 0.059      & 0.063      & 0.057      & 0.068      & 0.051      & 0.060     \\
    PalmTree               & 0.020      & 0.019      & 0.230      & 0.314      & \bf 0.878  & 0.292       & 0.006      & 0.007      & 0.080      & 0.184      & 0.676      & 0.191     \\
    Asm2Vec                & 0.494      & 0.460      & 0.444      & 0.535      & 0.563      & 0.499       & 0.290      & 0.252      & 0.234      & 0.343      & 0.376      & 0.299     \\
    CLAP                   & 0.244      & 0.221      & 0.214      & 0.550      & 0.781      & 0.402       & 0.187      & 0.176      & 0.168      & 0.455      & 0.707      & 0.339     \\ \hline
    Qwen 2.5 7B           & 0.471      & 0.412      & 0.343      & 0.456      & 0.608      & 0.458       & 0.342      & 0.301      & 0.234      & 0.345      & 0.488      & 0.342     \\
    Gemini 2.5 Flash       & \bf 0.739  & \bf 0.672  & \bf 0.568  & \bf 0.700  & 0.816      & \bf 0.699   & \bf 0.646  & \bf 0.579  & \bf 0.485  & \bf 0.618  & \bf 0.758  & \bf 0.617 \\ \hline \hline
    \end{tabular}
    }
    \caption{Evaluation of the baselines and our method on cross optimization retrieval with a pool size of \(1,000\).
    All functions are compiled for the \texttt{ARM} architecture using \texttt{gcc} with the optimization levels specified for each column.}
    \label{x-opt}
    \end{table*}
}

\begin{figure}[t]
\centerline{\includegraphics[width=\linewidth]{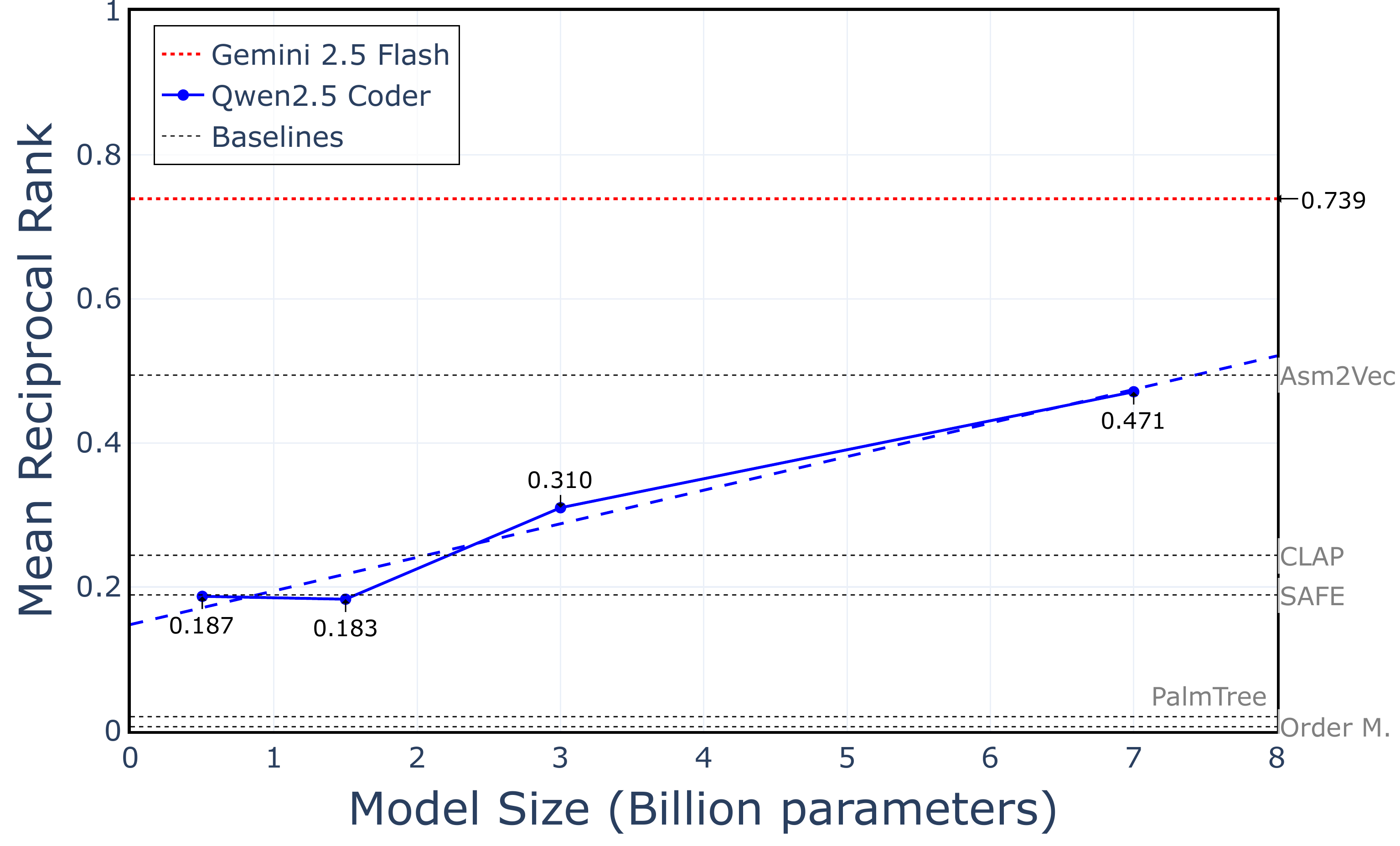}}
\caption{MRR for cross-optimization retrieval versus LLM parameters. 
All functions are compiled for \texttt{ARM}, with retrieval between O\(0\)--O\(1\) using a pool of \(1,000\) functions. 
Prompts include three few-shot examples.
}
\label{size-abl}
\end{figure}

In this experiment, we vary the language model size to determine the correlation between the number of parameters in the LLM and the performance
of our method on BCSD retrieval. Our results generally follow the scaling laws for neural language models ~\cite{scaling-laws}, in that increasing
the model size does significantly improve the results generated.

From our observations, LLMs with less than \(3\)B parameters do not seem to comprehend the analysis task when
they are not provided with any examples. When provided with examples, these small models will mimic the examples provided without basing the
output on the assembly function in the query. In ~\autoref{size-abl}, we observe a form of sub-linear increase in performance with respect to model size.
The Gemini 2.5 Flash model architecture is not disclosed at the time of writing, but we can expect the model to have at least an order
of magnitude more parameters than Qwen2.5 Code 7B, and may use a mixture-of-experts architecture, based on previous Gemini model architectures.

\subsection{Ablation Testing on Few-Shot Examples}

\begin{figure}
\centerline{\includegraphics[width=\linewidth]{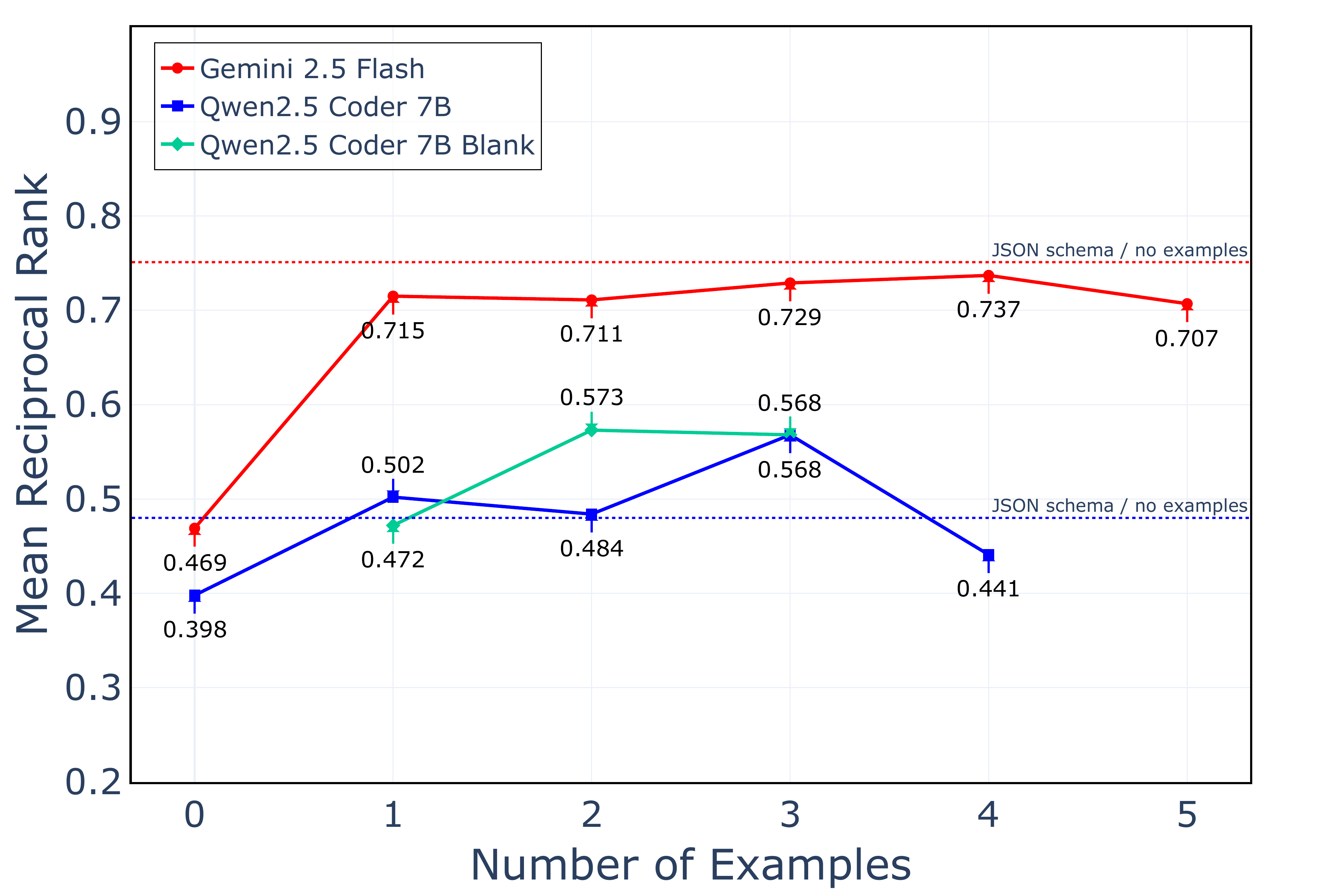}}
\caption{
MRR for cross-optimization retrieval versus prompt examples. 
Functions are compiled for \texttt{x86-64}, using a pool of \(100\) functions. 
Retrieval is between O\(0\)--O\(3\) for Gemini 2.5 Flash and O\(0\)--O\(1\) for Qwen2.5 Coder \(7\)B.
}
\label{ex-abl}
\end{figure}

By providing hand crafted examples to the large language model, we are able to increase the performance of assembly function analysis.
This follows the general observations behind few show prompting ~\cite{few-shot}. Providing a single example significantly increases the retrieval
scores, but providing more than one provides very limited increase in scores. Interestingly, a smaller model such as Qwen2.5 Coder \(7\)B still sees
marginal increase in MRR scores as the number of examples increases up to three. Another observation is the number of invalid \texttt{JSON} outputs generated by the local model. When no examples are provided, approximately \(3\%\) of queries generate an invalid output. With at least one example, or with a \texttt{JSON} schema, no invalid outputs are generated.

Another interesting result is the fact that when providing enough examples, the system prompt has very little impact on the retrieval scores.
This is depicted by the green data points in \autoref{ex-abl}, where only examples are provided with an empty string as the system prompt.
This suggests that the examples themselves contain enough information for the model to perform the analysis, without being explained exactly how.

With our experimental configuration, providing four examples to Qwen2.5 Coder \(7\)B significantly decreases its MRR scores. That is because the examples almost completely use the context window that we provide to the language model. As such, most assembly functions are too large to fit in the remaining tokens and are thus truncated, which loses information about our query.

The red dotted line in \autoref{ex-abl} represents the MRR score obtained by providing no examples to Gemini 2.5 Flash, but providing a reference \texttt{JSON} schema for the model to follow using the built-in capabilities of the API. The scores being very similar shows that Gemini does not base itself on the provided examples, but only uses them to understand the \texttt{JSON} schema required. In all our other evaluations, we provide examples instead of a \texttt{JSON} schema because local models do not have the capability of generating output based on a schema built-in. The blue dotted line in \autoref{ex-abl} simulates this behavior for Qwen2.5 Coder 7B by providing the schema within the system prompt. This does not match the results achieved with JSON-enforced token decoding, but almost matches the results obtained when a single example is provided.

\subsection{Ablation Testing on Prompt Features}

\begin{figure}
\centerline{\includegraphics[width=\linewidth]{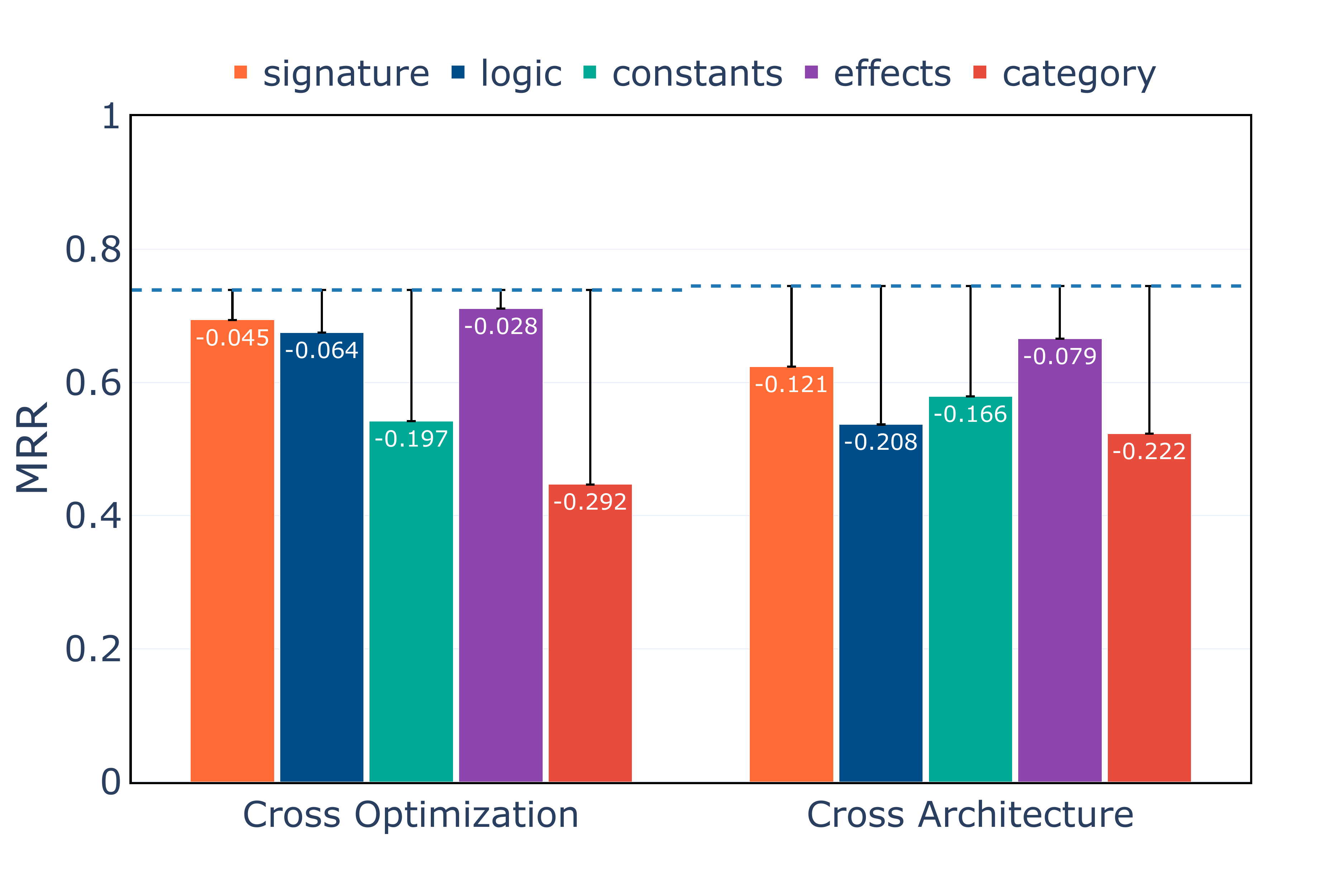}}
\caption{
MRR with one prompt section removed using Gemini 2.5 Flash. 
Cross-optimization retrieval is between O\(0\)--O\(3\) on \texttt{x86-64}, 
and cross-architecture retrieval is between \texttt{ARM} and \texttt{x86-64} 
(at O\(2\)). Both use a pool of \(100\) functions. 
Dotted lines indicate MRR without section removal; smaller bars imply greater impact.
}
\label{prompt-abl}
\end{figure}

To verify that each section of our prompt brings meaningful insight into the assembly function analysis task, we perform an ablation
study by removing one section of the prompt while keeping others intact. We notice that all individual sections bring a positive
outcome to the overall results, but some sections of the prompt have a larger impact than others. In particular, the categorization and
notable constants sections have the most impact on cross optimization retrieval. The categorization section causes an increase of almost
\(0.30\) on the MRR metric, when evaluated on the the hardest cross optimization task. The notable constants sections brings an increase of
almost \(0.20\) on the MRR for the same task. This result is justified by the comparatively large range of accepted values for both features.
For example, the list of notable constants has many more possible values and thus has more variability in the output compared to a set
of booleans, such as those in the side effects prompt section.

The story is slightly different when this ablation study is performed for cross architecture retrieval. As seen on \autoref{prompt-abl},
there is a smaller difference between the least and most influential prompt sections. The categorization and notable constants sections
are significantly less impactful than they were for the cross optimization study, while the other three sections have a larger impact.
For instance, the categorization section went from having an impact of \(0.30\) on the MRR to having an impact \(0.22\), while the logic
section is more influential, going from \(0.06\) to \(0.20\) in MRR impact.

\subsection{Combined Method}

Our method has shown strong performance in the BCSD retrieval tasks. Nevertheless, models trained for assembly function embedding
remain beneficial in specific subdomains, such as supporting niche architectures or understanding similarity through
specific obfuscation methods. State-of-the-art embedding models are much smaller than our method in terms of parameter count.
For instance, the CLAP ~\cite{CLAP} model baseline is only \(110\)M parameters, compared to the billions required to perform our method.
The cost of training such models is still very expensive,
but once trained, they are more lightweight and efficient than language models.

To achieve the best of both worlds, we experiment with combining the similarity scores from an embedding model and our prompting method.
Our evaluation consists of generating both an embedding and a LLM analysis for each function. The embedding similarity \(s_e\) is calculated
using cosine similarity, and the analysis similiarity \(s_a\) is calculated using jaccard similarity. Both similarity scores are then combined
with equal weight.
\[
    S = \frac{s_e + s_a}{2}
\]

In a real life scenario, combining both methods could be performed by using our textual representation as a pre-filter, and then calculating
the embedding representation. As such, the scalability and maintainability of the underlying BCSD database is maintained, while allowing a flexible
choice of embedding-based BCSD model. That is, an inverted index database is maintained for the pre-filter query. Once this query is completed, the
top-\(k\) results can be refined using an embedding model, by calculating the embeddings only for those \(k\) candidates. With this approach,
the decision to use an embedding model, and which one to use, can be made dynamically.

To provide a method that keeps all of the advantages of our presented work, we use Qwen3-Embedding \(4\)B ~\cite{qwen3} as the embedding model for this experiment.
As such, the combination still does not require any training nor fine-tuning. As shown in \autoref{composite}, off-the-shelf embedding models
based on a LLM perform very well. Furthermore, using a generic embedding model means that it can inexpensively be replaced by a new generation, since
the training phase is performed by the open source model developers.

\begin{table}
\centering
\footnotesize
\begin{tabular}{l|c|c|c}
\hline 
\hline 
                       & \makecell{Gemini\\2.5 Flash} 
                       & \makecell{Qwen\\Embedding} 
                       & \makecell{Combined\\(Both)} \\ \hline
\tt O0-O1           & 0.646 & 0.640 & \bf 0.910 \\
\tt O0-O2           & 0.579 & 0.554 & \bf 0.843 \\
\tt O0-O3           & 0.485 & 0.518 & \bf 0.759 \\
\tt O1-O3           & 0.618 & 0.640 & \bf 0.855 \\
\tt O2-O3           & 0.758 & 0.783 & \bf 0.921 \\
\tt ARM-AMD64       & 0.436 & 0.334 & \bf 0.736 \\
\tt PowerPC-AMD64   & 0.414 & 0.443 & \bf 0.746 \\
\tt MIPS-AMD64      & 0.417 & 0.415 & \bf 0.729 \\
\hline 
\hline 
\end{tabular}
\caption{Comparison between Qwen3-Embedding 4B, Gemini 2.5 Flash, and the combination of both. The retrieval task is performed on both cross
optimization and cross architecture settings. For cross optimization, the binaries are compiled for the \texttt{ARM} architecture, for cross architecture,
the binaries are compiled with optimization level \(2\). A pool of 1,000 assembly functions is used throughout. Only Recall@1 scores are presented.}
\label{composite}
\end{table}

The combined method significantly surpasses both the embedding and analysis methods individually, as well as the baseline methods. Seen differently, the embedding and analysis supplement
each other, meaning that our analysis extracts features from the assembly function that are not properly represented in the embedding model.

\section{Case Study}

\begin{figure}[t]
\centering
\setlength{\abovecaptionskip}{2pt}
\setlength{\belowcaptionskip}{2pt}
\setlength{\intextsep}{2pt}
\setlength{\tabcolsep}{3pt}
\renewcommand{\arraystretch}{0}

\begin{subfigure}{0.48\columnwidth} 
    \begin{minipage}{\linewidth}
    \begin{minted}[fontsize=\scriptsize,breaklines,frame=single]{nasm}
loc_21C58:
    PUSH {R4-R6,LR}
    MOV R5, R0
    ; ...
    MOV R0, #0x10624DD3
    LDR R2, [SP,#0x28+tp]
    ADD R6, SP, #0x28+tv
    SMULL.W R0, R1, R3, R0
    ; ...
    STM.W R5, {R0,R1}
    MOV R0, R5
    ADD SP, SP, #0x18
    POP {R4-R6,PC}
loc_21CA6:
    ; ...
loc_21CB2:
    BLX __stack_chk_fail
    \end{minted}
    \end{minipage}
    
    \vspace{5pt} 
    
    \begin{minipage}{\linewidth}
    \begin{minted}[fontsize=\scriptsize,breaklines,frame=single]{json}
{
  "operations": ["DataMovement", "SubroutineCall", "Arithmetic"],
  "category": "System/OS Interaction",
  "input_types": ["Pointer"],
  "modifies_inputs": true,
  "error_handling": true,
  "constants": ["0x28", "0x10624DD3", "0xAA260"],
  "return_type": "Pointer"
}
    \end{minted}
    \end{minipage}
\end{subfigure}
\hfill 
\begin{subfigure}{0.48\columnwidth}
    \begin{minipage}{\linewidth}
    \begin{minted}[fontsize=\scriptsize,breaklines,frame=single]{nasm}
loc_418210:
    ; ...
    MOV rdx, 20C49BA5E353F7CFh
    MOV [rsp+38h+var_38], rax
    MOV rax, rcx
    SAR rcx, 3Fh
    IMUL rdx
    ; ...
loc_41827E:
    ADD rsp, 38h
    RETN 
loc_418288:
    ; ...
loc_418294:
    CALL ___stack_chk_fail
    \end{minted}
    \end{minipage}
    
    \vspace{5pt} 
    
    \begin{minipage}{\linewidth}
    \begin{minted}[fontsize=\scriptsize,breaklines,frame=single]{json}
{
  "operations": ["DataMovement", "SubroutineCall", "Arithmetic"],
  "category": "System/OS Interaction",
  "input_types": [],
  "modifies_inputs": false,
  "error_handling": true,
  "constants": ["0x38", "0x20C49BA5E353F7CF", "0x28"],
  "return_type": "Integer"
}
    \end{minted}
    \end{minipage}
\end{subfigure}

\caption{\texttt{ARM} (left) and \texttt{x86-64} (right) assembly routines (top) with corresponding JSON extracted features (bottom).
The functions and feature sets are reduced to only show relevant
information for our case study.}
\label{fig:case-study}
\end{figure}

We perform a case study to better illustrate the capabilities and shortcomings of our method. We consider a simple example from a \texttt{ARM--x86-64} cross-architecture evaluation, with a \(1000\) function pool compiled with optimization level 3. Our method succeeds to match the \texttt{tvnow} function from the \texttt{curl} library. Our model ranked the correct match in first, with a similarity score of $0.559$. At a high level, the \texttt{tvnow} function retrieves the current time from the operating system. In its analysis, the LLM correctly identifies the high level properties, such as the ``System/OS Interaction" category, or the \texttt{subroutine\_call\_targets}. The model also accurately infers the behavior of the function:

\begin{tcolorbox}[colback=gray!10,colframe=gray!20,
 coltitle=black,]
\small
Time acquisition and conversion, potentially involving system calls to get current time and then performing a specific calculation (multiplication and shifting) on the nanoseconds component. It also includes stack cookie validation for security.
\end{tcolorbox}

In contrast, the identification of notable constants illustrates how our method can successfully extract syntactic elements, but emphasizes how static features do not generalize well. The analyses for both architectures correctly identify impactful constants, and also rightfully ignore irrelevant ones. The \texttt{x86-64} analysis identifies \texttt{0x20C49BA5E353F7CF} as notable, which is indeed used as part of the time conversion calculation. Precisely, this value is used as a fixed-point reciprocal multiplier to perform division by a constant divisor. The \texttt{ARM} analysis also correctly identifies the constant \texttt{0x10624DD3}, which is used for the same purpose. This illustrates a blind-spot in classical feature matching. Even though both analyses accurately capture the notable constants used in the function, the comparison mechanism failed because the compiler used semantically different instructions and thus different constants to perform the same operation.

The function signatures extracted in the feature set also differ slightly. The \texttt{x86-64} and \texttt{ARM} analyses disagree both on the input parameters of the function, and the return type. The \texttt{x86-64} analysis correctly identifies that \texttt{tvnow} does not have any input parameters, but does return an integer-like value. The \texttt{ARM} analysis does not agree on the return value, as it believes it to be a pointer. It also claims that the routine accepts a pointer as argument, which is incorrect from the point of view of the \texttt{curl} library source code. However, we observe in \autoref{fig:case-study} that the second instruction of the \texttt{arm} assembly routine does save \texttt{r0}, the first function argument in the ARM procedure call standard, into the variable register \texttt{r5}. This is most likely a case of return value optimization performed by the compiler. Towards the end of the function, the input now found in \texttt{r5} is used as an address to store the contents of the calculated time value to memory. This same address is then returned back to the caller, which is consistent with the analysis. Clearly, the model once again accurately identified the features in both analyses, but the model is not yet able to see through all compiler optimizations.


Even though this case study highlights a poor performance from our method, it still underscores the main value of our approach: \textit{interpretability}. The ability to inspect, understand, and verify the output generated by the model is novel, and allows for better human-model collaboration. After all, having a similarity model is unhelpful if the reverse engineer is not able to understand the reason for two functions having a high similarity score.

\section{Related Work}
\label{sec:related}


\noindent \textbf{Dynamic analysis.} This method family consists of analyzing the features of a binary or code fragment by monitoring its runtime behavior.
This method is compute intensive and requires a cross-platform emulator, but completely sidesteps the syntactic aspects of binary code
and solely analyzes its semantics. ~\cite{BCSD} As such, this method is highly resilient to obfuscations, but requires a sandboxed environment
and is hard to generalize across architectures and application binary interfaces ~\cite{blanket-exec, kam1n0}.

 \textbf{Code Representations.} Our research focuses on methods of binary similarity that use the assembly
code (and possibly the control flow graph) retrieved from the binary to perform similarity detection. However, other
approaches exist that use an alternate representation language, so that binaries compiled for all architectures are
extracted into the same representation language. Some methods turn to the well known LLVM-IR ~\cite{llvmir1, llvmir2, llvmir3}
as representation language, and others prefer the VEX intermediate representation ~\cite{VexIR2Vec, Zeek}.

\textbf{Full Program Similarity.} An alternate approach to binary matching is to compare whole binaries, instead
of their individual functions. This type of method constructs a fingerprint for the whole binary as compared to having one embedding
per function. This approach is primarily used for malware identification \cite{malware-id, malware-id2, malware-id3, malware-id4},
but also for identifying known vulnerabilities \cite{precise-static-vuln, firmware-vuln}.

\textbf{LLM-based software analysis}. Researchers have found many related applications of language models to binary and software analysis.
We give a sparse overview of the progress that has been made in this area of research.
For vulnerability detection, FuncVul ~\cite{funcVul} is a language-model-based method to detect whether a provided
source function is vulnerable. LLM4Decompile ~\cite{llm4decompile} tries to utilize LLMs for decompilation. An open source LLM is fine-tuned to learn the source code
representation of an assembly function, and the model is then evaluated on its decompilation capabilities. Fang et al. ~\cite{source-analysis}
measure the competency of LLMs for source code analysis, with a focus on obfuscated source code.
LLM-based fuzzing is another area that has recently gained interest. For instance, Asmita et al. ~\cite{llm-fuzz} use a LLM to
generate the initial seed of a fuzzing pipeline on the BusyBox ~\cite{busybox} executable.

\section{Conclusion}
\label{sec:conclusion}

Our method for BCSD is a shift in perspective compared to previous state-of-the-art BCSD embedding models. However,
this new approach maintains some of the known limitations and also brings new limitations. Most importantly, this
method makes use of massive models compared to previous methods. A powerful set of GPUs is required when generating
feature sets for a large pool or database. This favors centralised databases with large amounts of assembly fragments over
locally maintained databases with hundreds or thousands of functions. Otherwise, a commercially deployed LLM can be used at a
cost, but concerns surrounding data privacy can legitimately be raised.

We remain confident that our method brings significant improvements to the current landscape of BCSD, by resolving
many of the previously acknowledged limitations. Our method does not require any form of training, and can be performed with any LLM
available. It also offers a distinct advantage over the current state-of-the-art, because the feature vectors generated are human
interpretable. This makes the method easily tunable by human experts. The similarity scores are much more easily verifiable, and
cases of incorrect similarity detection can be explained using the generated feature set. Furthermore, this method can be scaled to
databases containing millions of assembly functions compared to embedding models, and remains accurate compared to approximated nearest neighbor search methods.

\bibliographystyle{plainurl}
\bibliography{references}


\end{document}